\documentclass[10pt,twocolumn,letterpaper]{article}

\usepackage{cvpr}              

\usepackage[dvipsnames]{xcolor}

\definecolor{cvprblue}{rgb}{0.21,0.49,0.74}
\usepackage[pagebackref,breaklinks,colorlinks,citecolor=cvprblue]{hyperref}

\def\paperID{*****} 
\def\confName{CVPR}
\def\confYear{2024}

\title{Tracking Small Birds by Detection Candidate Region Filtering and\\ Detection History-aware Association}

\author{Tingwei Liu\\
Nagoya University\\
Nagoya, Japan\\
{\tt\small liut@cs.is.i.nagoya-u.ac.jp}
\and
Yasutomo Kawanishi\\
RIKEN\\
Kyoto, Japan\\
{\tt\small yasutomo.kawanishi@riken.jp}
\and
Takahiro Komamizu\\
Nagoya University\\
Nagoya, Japan\\
{\tt\small taka-coma@acm.org}
\and
Ichiro Ide\\
Nagoya University\\
Nagoya, Japan\\
{\tt\small ide@i.nagoya-u.ac.jp}
}

\begin{document}
\maketitle
\begin{abstract}
This paper focuses on tracking birds that appear small in a panoramic video. When the size of the tracked object is small in the image (small object tracking) and move quickly, object detection and association suffers. To address these problems, we propose Adaptive Slicing Aided Hyper Inference (Adaptive SAHI), which reduces the candidate regions to apply detection, and Detection History-aware Similarity Criterion (DHSC), which accurately associates objects in consecutive frames based on the detection history. Experiments on the NUBird2022 dataset verifies the effectiveness of the proposed method by showing improvements in both accuracy and speed.
\end{abstract}    
\section{Introduction}
\label{sec:intro}

Observing bird behavior, especially movement among trees, is essential for understanding ecological activities, such as foraging and breeding. Although a variety of sensors are available for tracking birds, the study of bird behavior requires not only obtaining the location of the birds, but also the destination landscape and specific actions of the birds, which means it is best to analyze visual information. As a promising sensor, panoramic camera is preferable because it allows observation of a wide range of angle with a single device. 

\begin{figure}[t]
  \begin{center}
    \includegraphics[width = 1.0\columnwidth]{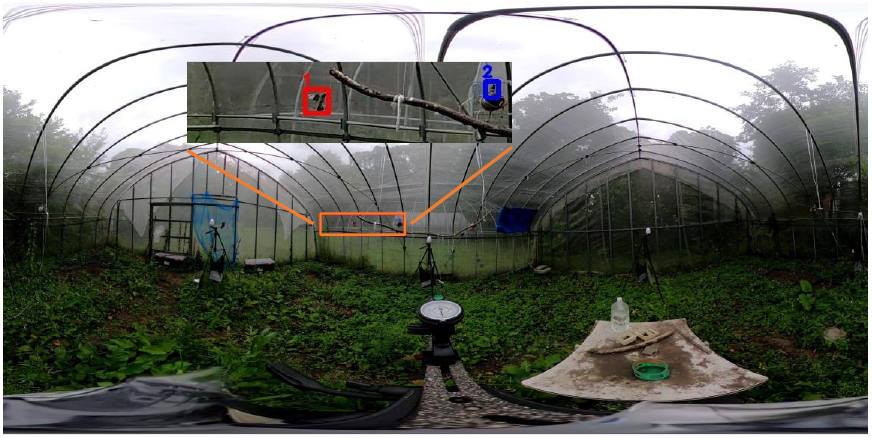}
  \end{center}
  \setlength{\abovecaptionskip}{-5pt}
  \setlength{\belowcaptionskip}{-15pt}
  \caption{Example of tracking results.}
  \label{fig:bird_tracking}
\end{figure}

However, there is a technical difficulty in detecting birds in panorama videos because they appear relatively smaller than taken from the same distance by a regular camera. In this paper, we tackle the problem of tracking birds that appear small in such videos. Figure ~\ref{fig:bird_tracking} shows an example of the tracking results, including the location (bounding box) and ID of the object.

In object tracking, Tracking-by-Detection (TbD) is the most common paradigm. It consists of two iterative processes: detecting objects from each frame, and associating them between frames. Due to the relatively small size of the target object (i.e., small object tracking), object detection and association suffers. First, regarding object detection, small object detection methods such as Slicing Aided Hyper Inference (SAHI)~\cite{SAHI} divide an image into small regions (detection candidate regions) and detect objects from each of them. However, only a small portion of the detection candidate regions actually contains birds, so these methods process many unnecessary regions. Meanwhile, for association, a common approach is to associate detected bounding boxes in a frame with bounding boxes predicted from the previous frame. In this process, the overlapping ratio of bounding boxes (Intersection-of-Union; IoU) and appearance feature similarity is often used as similarity metrics. However, because the birds in the images are small and move quickly, the detected bounding boxes often do not overlap even between adjacent frames. Also, it is not easy to extract appearance features sufficient to identify individual birds. Furthermore, when birds enter a nest or a birdhouse, long-time occlusion occurs, making time series predictions unreliable.

To address these problems, we propose Adaptive SAHI to filter detection candidate regions by extending SAHI~\cite{SAHI}, and Detection History-aware Similarity Criterion (DHSC) for association. We conduct experiments on the NUBird2022 dataset~\cite{Kawanishi2022detection,sumitani2021non} to show that the proposed method can improve the speed and accuracy of tracking.

\section{Related Work}
\label{sec:Related work}
\begin{figure}[tb]
    \begin{center}
        \includegraphics[width=1.0\linewidth]{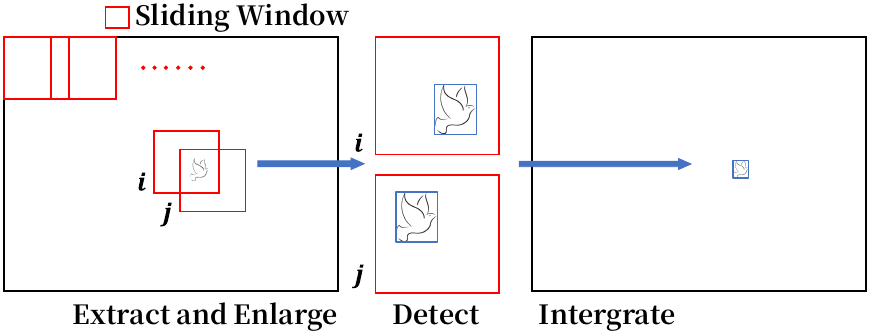}
    \end{center}
    \setlength{\abovecaptionskip}{-5pt}
    \setlength{\belowcaptionskip}{-15pt}
    \caption{Object detection with SAHI.}
    \label{fig:SAHI}
\end{figure}

\noindent\textbf{Small Object Detection}. Deep learning, which has become popular in recent years, has made it possible to detect objects that appear relatively large in images with high accuracy. However, due to the limitations of neural networks, it is required to scale the input image to a specific size (In general, smaller than the original image) before processing. In other words, objects in high-resolution will become very small after scaling. This problem is critical in small object detection tasks. Therefore, SAHI~\cite{SAHI}, specialized for small object detection, has been proposed. In SAHI, as shown in Figure~\ref{fig:SAHI}, the entire image is scanned using a small sliding window, allowing overlaps, to obtain detection candidate regions. Then, each of them is resized and input to an neural network-based object detector. Finally, all detection results are integrated in the coordinate system of the original image. Although this method is certainly effective for detecting small objects, the amount of computation increases depending on the number of detection candidate regions. Furthermore, when the number of small objects in an image is small, only a small portion of the regions actually contains the object, so, many unnecessary regions are processed. To address this problem, Kawanishi et al.~\cite{Kawanishi2022detection} proposed to use audio information to filter object detection candidate regions by sound source localization. Although it has shown improved performance, it is limited by the requirement for high-quality audio information, which is challenging to obtain in natural environments.

\noindent\textbf{Tracking-by-Detection (TbD)}. In TbD, objects in each frame are associated with those in the previous frames with respect to the similarity between bounding boxes. Simple Online and Real-time Tracking (SORT)~\cite{SORT}, a typical TbD method, uses a Kalman filter~\cite{kalman1960new} to estimate the state of an object (location and velocity) and associates the detected bounding boxes with predicted bounding boxes according to location-based similarity (i.e., IoU). However, since it is only based on location and velocity, if the object disappears (i.e., occlusion), errors will accumulate in the state estimation of that object, and when it reappears, it is likely to fail in association. To address this problem, DeepSORT~\cite{DeepSORT} introduces a similarity based on the appearance feature of objects to SORT and deals with the association failure by associating detections before and after occlusion. Furthermore, StrongSORT~\cite{StrongSORT} has updated each module of DeepSORT to improve object tracking performance by improving image feature extraction and restoring trajectories using Gaussian smooth interpolation.
Regarding the Kalman filter problem in occlusion, DeepSORT and StrongSORT utilize appearance features to realize the recovery of interrupted trajectories. However, in the case of small object tracking, the resolution of the object is very low, making it difficult to extract appearance features that can completely identify individuals.

\begin{figure*}[tb]
    \begin{center}
        \includegraphics[width=1.0\linewidth]{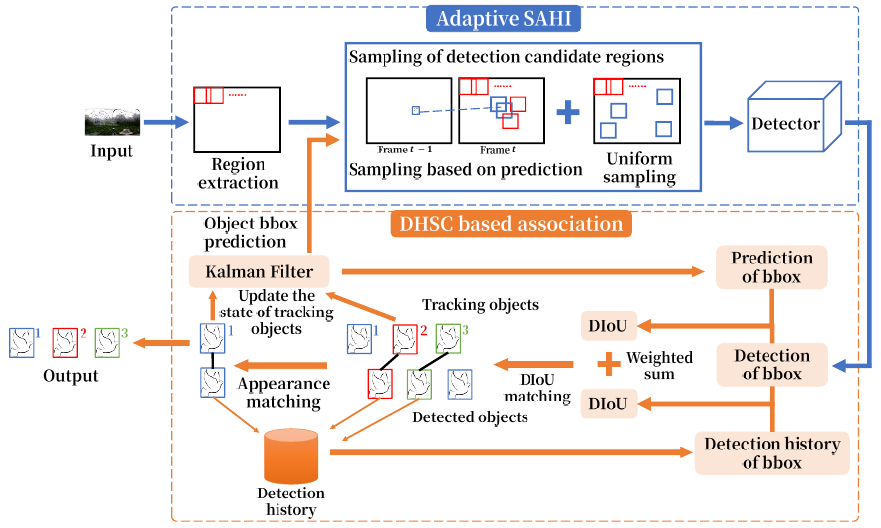}
    \end{center}
    \setlength{\abovecaptionskip}{-5pt}
    \setlength{\belowcaptionskip}{-15pt}
    \caption{Process flow of the proposed method.}
    \label{fig:framework}
\end{figure*}

\section{Proposed Method}
\label{sec:proposed method}
Figure~\ref{fig:framework} shows the process flow of the proposed method. The process consists of two parts: detection by Adaptive SAHI, and association based on DHSC.

\subsection{Detection by Adaptive SAHI}
\label{sec:Adaptive SAHI}

Since object tracking is a process between consecutive frames, there is a high possibility that the object to be tracked exists in the vicinity of the location detected in the previous frame. In object detection, according to SAHI~\cite{SAHI}, instead of directly using detection candidate regions generated by scanning the entire input image, the proposed method samples the detection candidate regions to run object detection efficiently. The proposed method reduces false positives as well as speeds up the process by sampling the detection candidate regions based on the tracking target locations. Specifically, for each tracking target, we first predict its location in the current frame, and then perform random sampling on the detection candidate regions containing the predicted location. In addition, in order to find tracking targets newly appeared in the scene, the proposed method additionally samples the detection candidate regions uniformly from the entire image without replacement. This is expected not only to find new objects, but also to rediscover objects whose tracking was interrupted due to occlusion. The final number of detection candidate regions is determined by the number of trajectories and the number of uniform samples. We name this sampling method adapted to tracking target locations as Adaptive SAHI.

\begin{figure}[t]
  \begin{center}
    \includegraphics[width = 1.0\columnwidth]{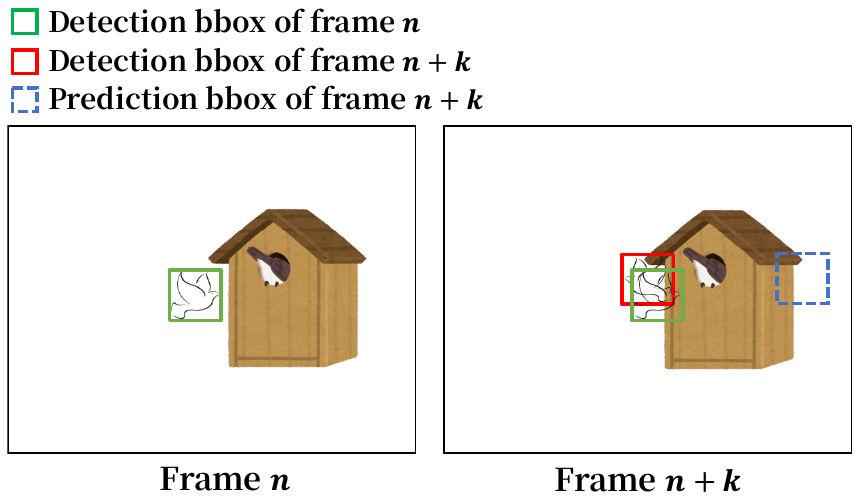}
  \end{center}
  \setlength{\abovecaptionskip}{-5pt}
  \setlength{\belowcaptionskip}{-15pt}
  \caption{Occlusion due to entry and exit.}
  \label{fig:Unreliable_situation}
\end{figure}

\subsection{Association based on DHSC}
\label{sec:DHSC}
Association in object tracking is to assign a consistent ID to each object by matching a previously tracked object with an object detected in the current frame according to the similarity. As mentioned in Section ~\ref{sec:Related work}, appearance features cannot completely identify individual birds. However, the location-based similarity is easy to fail when different objects move across each other, or the locations where a single object appears before and after are too far apart, so we follow a two-stage matching process. The first stage is based on location and the second stage is based on appearance. For associating the birds that are small and move quickly, Distance-IoU (DIoU)~\cite{DIoU} is used in the first stage. The second stage of matching is performed only for objects that could not be matched in the first stage of matching. It uses cosine similarity between appearance features. 

Occlusion due to entering and exiting a shelter (i.e., nest or birdhouse) with a single entrance/exit as shown in Figure~\ref{fig:Unreliable_situation} often occurs in bird observation considered in this study. If a bird enters a birdhouse in frame $n$, the Kalman filter continues to make predictions based on the direction and movement before it disappears. At frame $n+k$, when the bird reappears from the birdhouse, it predicts the bird at the location indicated by the blue dashed line in the upper right corner of the birdhouse, which causes a mistake in the association. Thus, predictions of the Kalman filter is sometimes unreliable when a long-time occlusion occurs. In fact, since it is an occlusion due to entry and exit, the bird is at the same place when it reappears as when it disappeared, so it can be correlated with the detection history. Therefore, the proposed method saves the latest detection result of each tracked object as its detection history, and uses it and predicted locations for similarity calculation with current detection results. By calculating the weighted sum of the similarities, the similarity calculation becomes effective even when predictions are unreliable, and accuracy is expected to improve. Note that the weight of detection histories and predicted locations is determined experimentally.
\section{Experiment}
\subsection{Experiment Setting}
\noindent\textbf{Baseline}. As the baseline method for this experiment, we use You Only Look Once (YOLO) v5~\cite{yolov5} with SAHI as a detector, and StrongSORT~\cite{StrongSORT} as a tracker.

\noindent\textbf{Dataset}. We use a dataset called NUBird2022~\cite{Kawanishi2022detection,sumitani2021non}\footnote{\url{https://www.cs.is.i.nagoya-u.ac.jp/opensource/nubird/}(Accessed May 23, 2024)}, which consists of approximately 8 minutes of video of a closed environment where five Zebra finches (\textit{Taeniopygia guttata}) are kept. The length of the video is 14,459 frames, and the resolution of each frame is $4,096\times2,048$ pixels. The size of the birds in the image is only about $10\times20$ pixels, and their flying speed can reach 20 pixels per frame.

\begin{table}[tb]
    \begin{center}
    \setlength{\abovecaptionskip}{5pt}
    \caption{Comparison of the performance of the baseline and the proposed methods in the evaluation set.}\label{table:experiment result}
    \begin{subtable}[h]{0.47\textwidth}
        \centering
            \caption{Scene 1}
            \resizebox{\linewidth}{!}{
            \begin{tabular}{ccccccc}
                \toprule
                \multicolumn{1}{c}{Method}& \multicolumn{1}{|c}{MOTA$\uparrow$}& IDF$_1$$\uparrow$& FP$\downarrow$& FN$\downarrow$& IDsw$\downarrow$& \multicolumn{1}{c}{FPS$\uparrow$}\\
                \midrule
                Baseline& \multicolumn{1}{|r}{$-119.6$}& \multicolumn{1}{r}{20.2}& \multicolumn{1}{r}{3,049}& \multicolumn{1}{r}{\textbf{697}}& \multicolumn{1}{r}{91}& \multicolumn{1}{r}{0.2}\\
                Proposed& \multicolumn{1}{|r}{\textbf{24.2}}& \multicolumn{1}{r}{\textbf{27.0}}& \multicolumn{1}{r}{\textbf{591}}& \multicolumn{1}{r}{716}& \multicolumn{1}{r}{\textbf{16}}& \multicolumn{1}{r}{\textbf{4.4}}\\
                \bottomrule
            \end{tabular}
            }
    \end{subtable}\\
    
    \begin{subtable}[h]{0.47\textwidth}
        \centering
            \setlength{\belowcaptionskip}{5pt} 
            \caption{Scene 2}
            \resizebox{\linewidth}{!}{
            \begin{tabular}{ccccccc}
                \toprule
                \multicolumn{1}{c}{Method}& \multicolumn{1}{|c}{MOTA$\uparrow$}& IDF$_1$$\uparrow$& FP$\downarrow$& FN$\downarrow$& IDsw$\downarrow$& \multicolumn{1}{c}{FPS$\uparrow$}\\
                \midrule
                Baseline& \multicolumn{1}{|r}{$-64.5$}& \multicolumn{1}{r}{22.3}& \multicolumn{1}{r}{3,435}& \multicolumn{1}{r}{644}& \multicolumn{1}{r}{150}& \multicolumn{1}{r}{0.2}\\
                Proposed& \multicolumn{1}{|r}{\textbf{71.5}}& \multicolumn{1}{r}{\textbf{50.2}}& \multicolumn{1}{r}{\textbf{205}}& \multicolumn{1}{r}{\textbf{500}}& \multicolumn{1}{r}{\textbf{28}}& \multicolumn{1}{r}{\textbf{8.0}}\\
                \bottomrule
            \end{tabular}
            }
    \end{subtable}
    \end{center}
\vspace{-10pt}
\end{table}

\begin{figure}[tb]
  \begin{center}
    \subfloat[Scene 1]{\includegraphics[width=.49\columnwidth]{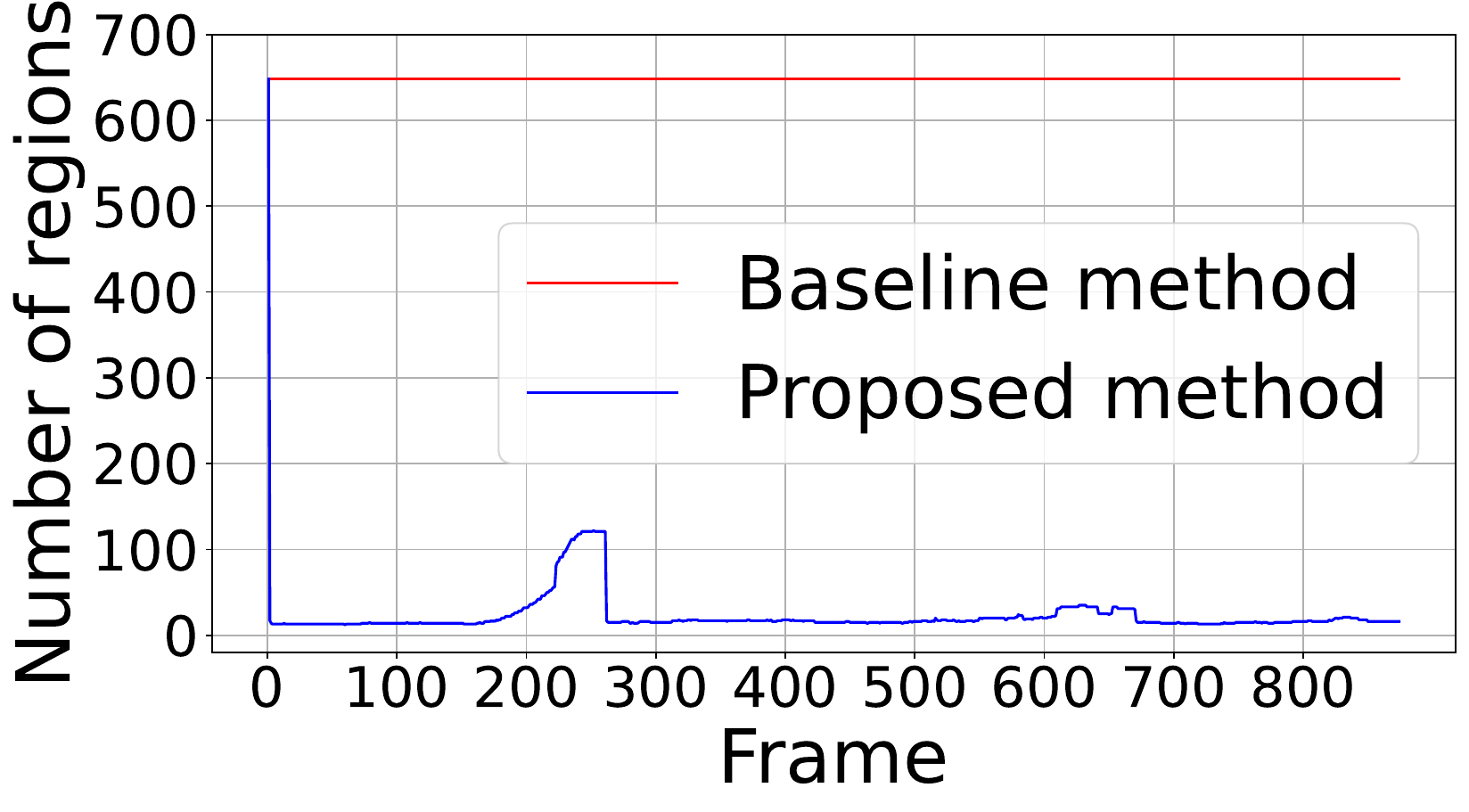}}\hspace{2pt}
	\subfloat[Scene 2]{\includegraphics[width=.49\columnwidth]{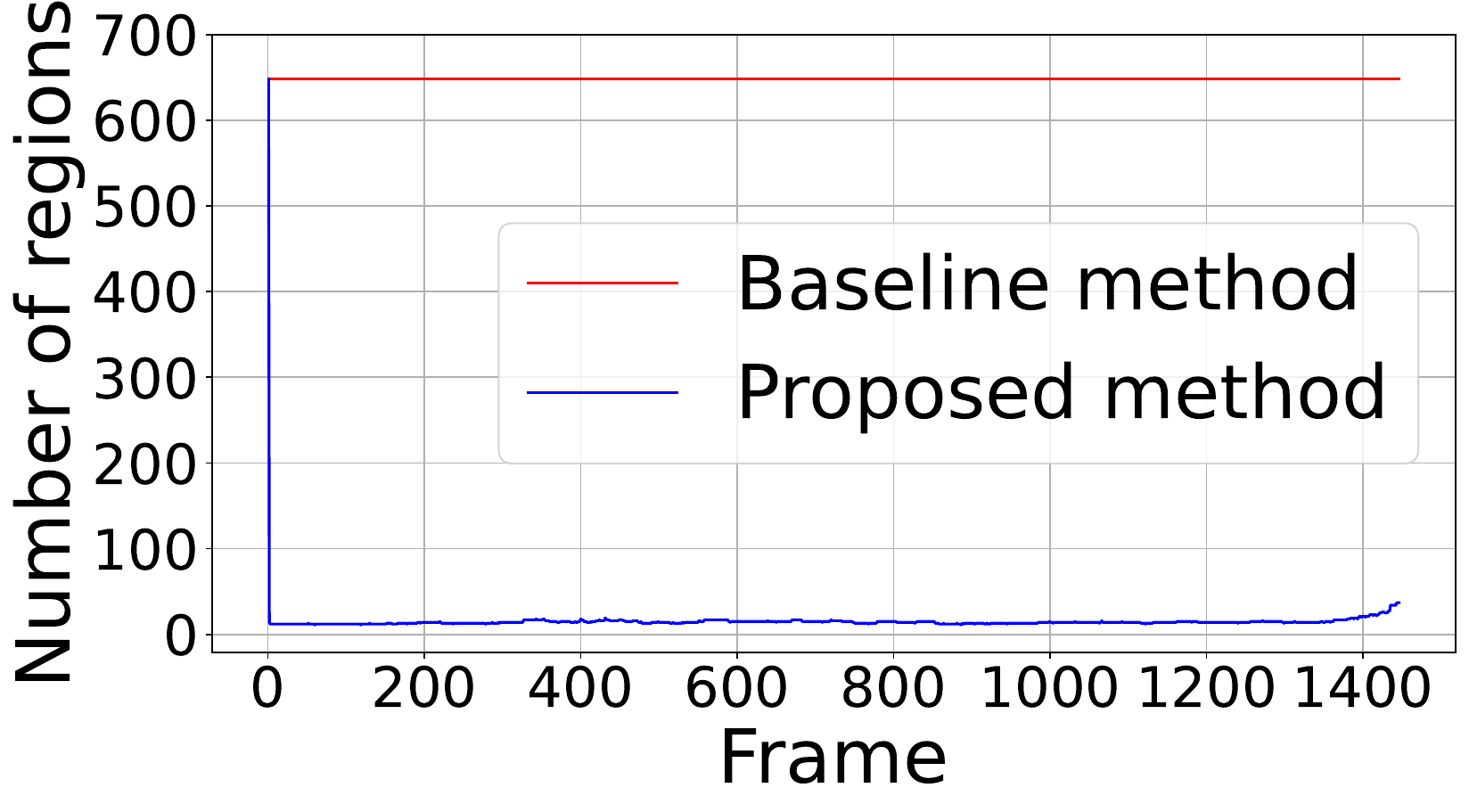}}
  \end{center}
  \setlength{\abovecaptionskip}{-5pt}
  \setlength{\belowcaptionskip}{-15pt}
  \caption{Change in number of detection candidate regions.}
  \label{fig:regions change}
\end{figure}

Two scenes containing nonlinear movement of birds and occlusions are selected from the original video to serve as the evaluation set, and the remaining images are randomly assigned to the training set and validation set for training the detector at a ratio of $3:1$. Slicing aided Fine-tuning (SF)~\cite{SAHI} is applied to the training and validation sets to augment the dataset. Specifically, for both sets of images, the entire image is scanned using a small sliding window, allowing for overlap, and the region containing the bird is extracted, scaled, and added to the original set. For images containing birds in training and validation sets, regions where at least one bird exists are extracted and added to the corresponding set for augmentation. For images with no bird, all regions are included in the training set for augmentation. The size of the sliding window is $256\times128$ pixels, the overlap rate is 0.25, and the resolution after scaling is $640\times320$ pixels. The results after the augmentation are as follows.
\begin{itemize}
    \item Detector training set: 27,395 images, of which 9,095 are original images (1 background image) and 18,300 are augmented images (441 background images). 
    \item Detector validation set: 9,157 images, of which 3,039 are original images and 6,118 are augmented images.
    \item Evaluation set:
    \begin{itemize}
        \item Scene 1: 874 frames. A scene where birds take continuous cross-movement.
        \item Scene 2: 1,451 frames. A scene where two birds alternately enter and exit the entrance of a birdhouse.
    \end{itemize}
\end{itemize}

\noindent\textbf{Metrics}. The evaluation metrics used in this experiment are False Positive (FP), False Negative (FN), IDentity switches (IDsw)~\cite{IDsw}, Multi-Object Tracking Accuracy (MOTA)~\cite{MOTA}, IDF$_1$ Score (IDF$_1$)~\cite{IDF1}, and processed Frames Per Second (FPS).

\begin{figure}[t]
  \begin{center}
    \includegraphics[width = 1.0\columnwidth]{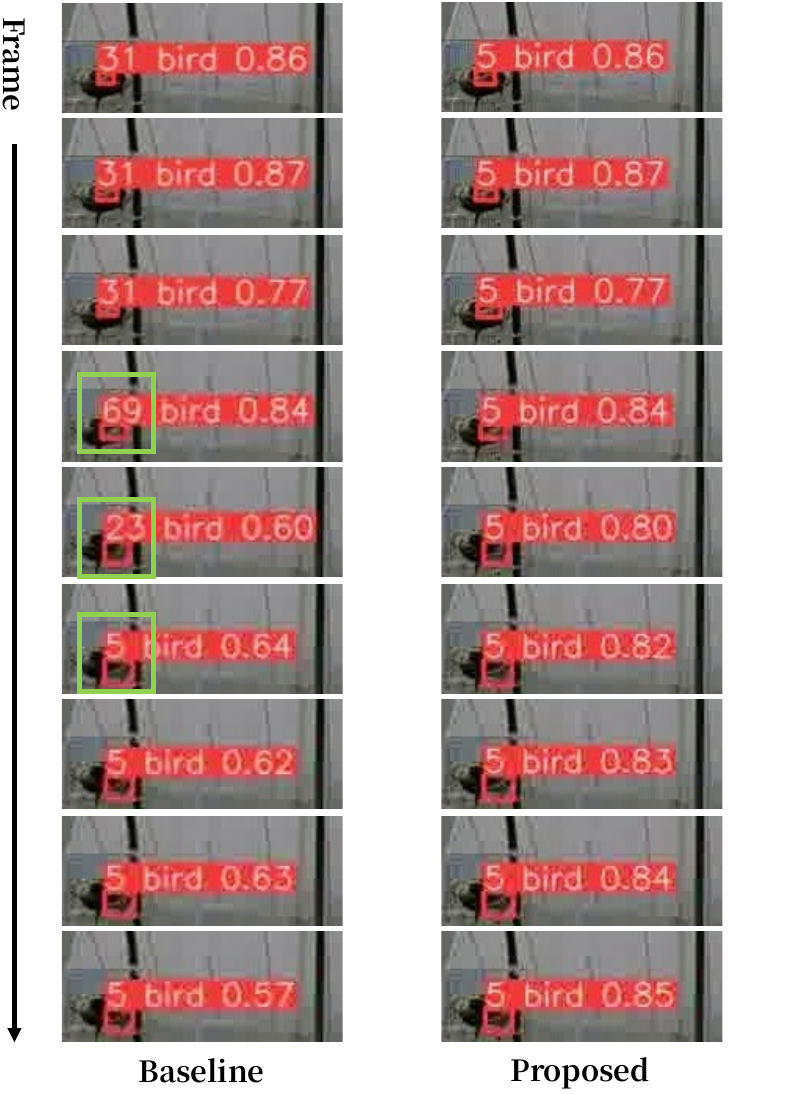}
  \end{center}
  \setlength{\abovecaptionskip}{-5pt}
  \setlength{\belowcaptionskip}{-15pt}
  \caption{Visualization of tracking results.}
  \label{fig:visualization of tracking}
\end{figure}

\subsection{Experiment Result}
Table~\ref{table:experiment result} shows the experiment results in the evaluation set. MOTA improved by 143.8 points in Scene~1 and 136.0 points in Scene~2. On the other hand, IDF$_1$ improved by 6.8 percent points in Scene~1 and 27.9 percent points in Scene~2. FPS improved by 4.2 in Scene~1 and 7.8 in Scene~2. FP decreased by 2,458 in Scene~1 and 3,230 in Scene~2. IDsw decreased by 75 times in Scene~1 and 122 times in Scene~2. Figure~\ref{fig:regions change} shows the change in the number of detection candidate regions. Although the number of detection candidate regions increased due to the increase in erroneous trajectories in Scene~1, the proposed method reduced them by 96.5\% in Scene~1 and 97.7\% in Scene~2.

Figure~\ref{fig:visualization of tracking} shows a portion of the tracking results when a bird reappears from the birdhouse in Scene 2. We can see that the proposed method could track the bird more stably compared to the baseline method. In the nine frames of tracking, the baseline method experienced IDsw three times and eventually continued tracking with an ID different from the initial one. In contrast, the proposed method maintained stable tracking throughout the nine frames. This demonstrates the effectiveness of the proposed method in handling such occlusions.

\subsection{Ablation Study}
\subsubsection{Evaluation of Similarity Metrics}
Various combinations of similarity metrics were tested on Scene 2 where occlusion occurred. To remove the influence from detection, we used the ground truth for the detection bounding box. Results are shown in Table~\ref{table:similarity experiment result}, where DH-DIoU represents the proposed DHSC using DIoU as the similarity metric.

First, each two-stage matching method showed the same or higher accuracy than the corresponding one-stage matching method, confirming that two-stage matching is useful for improving accuracy. When appearance features were used in the first stage, many ID switches occurred. The reason for this is that it was difficult to extract discriminative appearance features of birds in the scene. Next, focusing on the differences between the second and third lines, the fourth and fifth lines, and the sixth and seventh lines, we can see that changing DIoU to DH-DIoU increased IDF$_1$ by 0.8, 4.8, and 9.8 percent points, respectively. The combination of appearance features and DH-DIoU obtained the highest IDF$_1$, but when evaluated comprehensively, DH-DIoU and appearance features were better. From these results, we confirmed the effectiveness of the proposed method, which used DH-DIoU in the first stage and appearance features in the second stage.

\begin{table}[tb]
\centering
\setlength{\abovecaptionskip}{5pt}
\caption{Comparison of different similarity metrics.}
\resizebox{0.94\linewidth}{!}{
\begin{tabular}{ccccc}
\toprule
\multicolumn{1}{c}{First stage}& \multicolumn{1}{c}{Second stage}& \multicolumn{1}{|c}{MOTA$\uparrow$}& IDF$_1$$\uparrow$& IDsw$\downarrow$\\
\midrule
Appearance&\multicolumn{1}{c}{---}& \multicolumn{1}{|r}{81.8}& \multicolumn{1}{r}{30.7}& \multicolumn{1}{r}{176}\\
DIoU&\multicolumn{1}{c}{---}& \multicolumn{1}{|r}{\textbf{99.0}}& \multicolumn{1}{r}{40.6}& \multicolumn{1}{r}{8}\\
DH-DIoU&\multicolumn{1}{c}{---}& \multicolumn{1}{|r}{\textbf{99.0}}& \multicolumn{1}{r}{41.4}& \multicolumn{1}{r}{\underline{6}}\\
Appearance & DIoU& \multicolumn{1}{|r}{90.0}& \multicolumn{1}{r}{46.4}& \multicolumn{1}{r}{150}\\
Appearance & DH-DIoU& \multicolumn{1}{|r}{\underline{96.7}}& \multicolumn{1}{r}{\textbf{51.2}}& \multicolumn{1}{r}{51}\\
DIoU & Appearance& \multicolumn{1}{|r}{\textbf{99.0}}& \multicolumn{1}{r}{40.7}& \multicolumn{1}{r}{8}\\
DH-DIoU & Appearance& \multicolumn{1}{|r}{\textbf{99.0}}& \multicolumn{1}{r}{\underline{50.5}}& \multicolumn{1}{r}{\textbf{5}}\\
\bottomrule
\end{tabular}
}
\label{table:similarity experiment result}
\vspace{-10pt}
\end{table}

\begin{table}[tb]
\footnotesize
\centering
\setlength{\abovecaptionskip}{5pt}
\caption{Comparison of different weights.}
\resizebox{0.94\linewidth}{!}{
\begin{tabular}{ccccc}
\toprule
\multicolumn{1}{c}{Predictions}& \multicolumn{1}{c}{Histories}& \multicolumn{1}{|c}{MOTA$\uparrow$}& IDF$_1$$\uparrow$& IDsw$\downarrow$\\
\midrule
0.1& 0.9& \multicolumn{1}{|r}{\underline{69.4}}& \multicolumn{1}{r}{49.8}& \multicolumn{1}{r}{31}\\
0.2& 0.8& \multicolumn{1}{|r}{\textbf{71.5}}& \multicolumn{1}{r}{\underline{50.2}}& \multicolumn{1}{r}{\textbf{28}}\\
0.3& 0.7& \multicolumn{1}{|r}{59.1}& \multicolumn{1}{r}{48.7}& \multicolumn{1}{r}{31}\\
0.4& 0.6& \multicolumn{1}{|r}{68.5}& \multicolumn{1}{r}{\underline{50.2}}& \multicolumn{1}{r}{41}\\
0.5& 0.5& \multicolumn{1}{|r}{68.3}& \multicolumn{1}{r}{\textbf{54.3}}& \multicolumn{1}{r}{\underline{29}}\\
0.6& 0.4& \multicolumn{1}{|r}{32.8}& \multicolumn{1}{r}{29.9}& \multicolumn{1}{r}{47}\\
0.7& 0.3& \multicolumn{1}{|r}{38.1}& \multicolumn{1}{r}{39.5}& \multicolumn{1}{r}{34}\\
0.8& 0.2& \multicolumn{1}{|r}{31.2}& \multicolumn{1}{r}{31.2}& \multicolumn{1}{r}{33}\\
0.9& 0.1& \multicolumn{1}{|r}{54.0}& \multicolumn{1}{r}{40.7}& \multicolumn{1}{r}{33}\\
\bottomrule
\end{tabular}
}
\label{table:weights experiment result}
\vspace{-10pt}
\end{table}

\subsubsection{Evaluation of Different Weights}
To determine the weights of predicted locations and detection histories in DHSC, different weights were tested on Scene 2. Results are shown in Table~\ref{table:weights experiment result}. We can see that when the weight of detection histories was below 0.5, the scores of the metrics were significantly lower than when the weight was 0.5 or higher. This indicates that detection histories play a crucial role in handling occlusions in Scene 2. Considering the three metrics, we ultimately selected (0.2, 0.8) as the weights for our experiments.

\section{Conclusion}
This paper proposed Adaptive SAHI for filtering the detection candidate regions and DHSC for stable association when occlusion occurs. Experiment on NUBird2022~\cite{Kawanishi2022detection,sumitani2021non} confirmed the effectiveness of the proposed method. As future work, we are considering applying the proposed methods to other models and multimodal tracking.

\section*{Acknowledgement}
This paper was partly supported by JSPS KAKENHI Grant Numbers JP20H00475 and JP21H03519.
{
    \small
    \bibliographystyle{ieeenat_fullname}
    \bibliography{main}

\begin{thebibliography}{12}
\providecommand{\natexlab}[1]{#1}
\providecommand{\url}[1]{\texttt{#1}}
\expandafter\ifx\csname urlstyle\endcsname\relax
  \providecommand{\doi}[1]{doi: #1}\else
  \providecommand{\doi}{doi: \begingroup \urlstyle{rm}\Url}\fi

\bibitem[Akyon et~al.(2022)Akyon, Altinuc, and Temizel]{SAHI}
Fatih~Cagatay Akyon, Sinan~Onur Altinuc, and Alptekin Temizel.
\newblock Slicing aided hyper inference and fine-tuning for small object detection.
\newblock In \emph{Proceedings of the 2022 IEEE International Conference on Image Processing}, pages 966--970, 2022.

\bibitem[Bernardin and Stiefelhagen(2008)]{MOTA}
Keni Bernardin and Rainer Stiefelhagen.
\newblock Evaluating multiple object tracking performance: {The CLEAR MOT metrics}.
\newblock \emph{EURASIP Journal on Image and Video Processing}, 2008\penalty0 (246309):\penalty0 1--10, 2008.

\bibitem[Bewley et~al.(2016)Bewley, Ge, Ott, Ramos, and Upcroft]{SORT}
Alex Bewley, Zongyuan Ge, Lionel Ott, Fabio Ramos, and Ben Upcroft.
\newblock Simple online and realtime tracking.
\newblock In \emph{Proceedings of the 2016 IEEE International Conference on Image Processing}, pages 3464--3468, 2016.

\bibitem[Du et~al.(2023)Du, Zhao, Song, Zhao, Su, Gong, and Meng]{StrongSORT}
Yunhao Du, Zhicheng Zhao, Yang Song, Yanyun Zhao, Fei Su, Tao Gong, and Hongying Meng.
\newblock Strong{SORT}: Make {DeepSORT} great again.
\newblock \emph{IEEE Transactions on Multimedia}, 25:\penalty0 8725--8737, 2023.

\bibitem[Jocher(2020. https://github.com/ultralytics/yolov5/ (Accessed May 23, 2024))]{yolov5}
Glenn Jocher.
\newblock Ultralytics {YOLOv5}, 2020. https://github.com/ultralytics/yolov5/ (Accessed May 23, 2024).

\bibitem[Kalman(1960)]{kalman1960new}
Rudolph~Emil Kalman.
\newblock A new approach to linear filtering and prediction problems.
\newblock \emph{Journal of Basic Engineering}, 82\penalty0 (1):\penalty0 35--45, 1960.

\bibitem[Kawanishi et~al.(2022)Kawanishi, Ide, Chu, Matsuhira, Kastner, Komamizu, and Deguchi]{Kawanishi2022detection}
Yasutomo Kawanishi, Ichiro Ide, Baidong Chu, Chihaya Matsuhira, Marc~A Kastner, Takahiro Komamizu, and Daisuke Deguchi.
\newblock Detection of birds in a {3D} environment referring to audio-visual information.
\newblock In \emph{Proceedings of the 18th IEEE International Conference on Advanced Video and Signal Based Surveillance}, pages 1--7, 2022.

\bibitem[Li et~al.(2009)Li, Huang, and Nevatia]{IDsw}
Yuan Li, Chang Huang, and Ram Nevatia.
\newblock Learning to associate: {HybridBoosted} multi-target tracker for crowded scene.
\newblock In \emph{Proceedings of the 2009 IEEE Computer Society Conference on Computer Vision and Pattern Recognition}, pages 2953--2960, 2009.

\bibitem[Ristani et~al.(2016)Ristani, Solera, Zou, Cucchiara, and Tomasi]{IDF1}
Ergys Ristani, Francesco Solera, Roger Zou, Rita Cucchiara, and Carlo Tomasi.
\newblock Performance measures and a data set for multi-target, multi-camera tracking.
\newblock In \emph{Computer Vision {---}ECCV 2016 Workshops: Amsterdam, The Netherlands, October 8--10 and 15--16, 2016, Proceedings, Part II, Lecture Notes in Computer Science}, vol.9914, pages 17--35. Springer, 2016.

\bibitem[Sumitani et~al.(2021)Sumitani, Suzuki, Arita, Nakadai, and Okuno]{sumitani2021non}
Shinji Sumitani, Reiji Suzuki, Takaya Arita, Kazuhiro Nakadai, and Hiroshi~G Okuno.
\newblock Non-invasive monitoring of the spatio-temporal dynamics of vocalizations among songbirds in a semi free-flight environment using robot audition techniques.
\newblock \emph{Birds}, 2\penalty0 (2):\penalty0 158--172, 2021.

\bibitem[Wojke et~al.(2017)Wojke, Bewley, and Paulus]{DeepSORT}
Nicolai Wojke, Alex Bewley, and Dietrich Paulus.
\newblock Simple online and realtime tracking with a deep association metric.
\newblock In \emph{Proceedings of the 2017 IEEE International Conference on Image Processing}, pages 3645--3649, 2017.

\bibitem[Zheng et~al.(2020)Zheng, Wang, Liu, Li, Ye, and Ren]{DIoU}
Zhaohui Zheng, Ping Wang, Wei Liu, Jinze Li, Rongguang Ye, and Dongwei Ren.
\newblock Distance-{IoU} loss: Faster and better learning for bounding box regression.
\newblock In \emph{Proceedings of the 34th AAAI Conference on Artificial Intelligence}, pages 12993--13000, 2020.

\end{thebibliography}
}


\end{document}